\documentclass[letterpaper, 10 pt, conference]{ieeeconf}  

\IEEEoverridecommandlockouts                              

\usepackage[nolist]{acronym}
\usepackage{cite}
\usepackage{color, soul}
\usepackage{hyperref}
\usepackage{amsmath}
\usepackage{booktabs}
\usepackage{tabularx}
\usepackage{makecell}

\usepackage{xcolor}
\usepackage{lipsum}
\usepackage{tcolorbox}

\title{\LARGE \bf
Mitigating Concept Drift in QoS Prediction for Teleoperation of Autonomous Vehicles Using Historic Data
}

\author{Xiyan Su$^{*}$$^{1}$, Jianning Gao$^{*}$, Mahmoud Ashri$^{*}$, and Frank Diermeyer$^{*}$
    \thanks{The research is funded by the Federal Ministry of Education and Research of Germany (BMBF) within the project ConnRAD (FKZ 16KISR034),}
    \thanks{$^{*}$The authors are with the Institute of Automotive Technology, Technical University of Munich (TUM), Boltzmannstr.15, DE-85748, Garching bei München, Germany.}%
    \thanks{$^{1}$ Corresponding author {\tt\small xiyan.su@tum.de}}%
}

\begin{document}

\maketitle
\thispagestyle{empty}
\pagestyle{empty}

\begin{abstract}



Teleoperation serves as the fallback solution to autonomous driving but reliable functions of the teleoperation require a certain amount of mobile network resources, which cannot be guaranteed at all times. Therefore, predictive quality of service (pQoS) is introduced as a concept to increase the resilience of the teleoperation. In this paper, based on a data measurement campaign, we propose a prediction framework to prediction two important network KPIs of teleoperation: uplink data-rate and round-trip latency. Furthermore, we introduce a method to alleviate the performance degradation of machine-learning-based prediction models on previously unseen data due to concept drift by incorporating historic data into the prediction pipeline. Additionally, we introduce the metric of critical scenario detection to evaluate the prediction performance specifically for teleoperation. 



\end{abstract}

\section{INTRODUCTION}
\label{sec:introduction}

\subsection{Teleoperation}

Teleoperation serves as the fallback solution in the level-4 \ac{ad} defined by SAE \cite{SAEInternational2021}. In teleoperation, the vehicle sends sensor data to the operator and the operator sends control commands to the vehicle over mobile network. The mobile network is proven to be sufficient for teleoperation in general \cite{Neumeier2019}. However, reliable functions require a certain amount of network resources, which the mobile network is not able to provide at all times currently. For example, \ac{5gaa} defines in its technical report that it requires 32 Mbps \ac{ul} data-rate and 99\% reliability for a safe teleoperation scenario \cite{5GAA2020}. However, the current measured commercial mobile network cannot provide such resources at all times. 


In the latest version of \cite{TUMFTM2025}, some efforts have gone into reducing the data-rate usage of teleoperation, such as using the state-of-the-art video codec to reduce the data-rate of the video streams while maintaining the same video quality, and using multiple links to increase the available bandwidth. However, these efforts still cannot guarantee a safe operation through the network at all times. It is worth noticing that, in most critical scenarios, the vehicle can still be teleoperated with degraded quality, e.g., with lower video quality due to limited \ac{ul} data-rate, or with velocity restrictions due to high speed latency \cite{Kulzer2021}.

\subsection{Predictive Quality of Service}

In this context, the concept of \ac{pqos} is put forward by \ac{5gaa} \cite{5GAA2019}. \ac{pqos} allows \ac{ad} vehicles to predict the upcoming network quality and adapt to it proactively. This particularly increases the resilience of the teleoperation against the fluctuating mobile network quality. 

As described in \cite{5GAA2019}, \ac{pqos} can be deployed either at the \ac{mno}, or be provided as an \ac{ott} solution by a third party. The 3rd Generation Partnership Project also introduced the Network Data Analytics Function that defines the data exchange between different parties to enable the functionality of \ac{pqos} \cite{ETSI2024}. However, the functionality of \ac{pqos} is still not supported ubiquitously by the \ac{mno}. Therefore, we focus on the \ac{ott} deployment in this paper, i.e., the \ac{pqos} is provided as a service on the vehicle by a third-party service provider, e.g., vehicle manufacturer.

\subsection{Related Works}
\label{sec:related_works}

Predicting the \ac{qos} of mobile network per se is a complex problem, as it is affected by different aspects of the network, e.g., \ac{ran}, core network, and routing. Some works have gone into analyzing the \ac{qos} data. The results in \cite{Akselrod2017} show the temporal and spatial variance of the \ac{lte} \ac{dl} data-rate as well as the influence of signal quality and main transport protocol on it. In \cite{Ide2016}, it discusses the cause of the asymmetry of modern \ac{lte} technology in detail and shows that the \ac{dl} channel indicators, such as \ac{rsrp}, \ac{sinr}, and \ac{rsrq} are suitable indicators for \ac{ul} performance. Therefore, the similar \ac{phy} \acp{kpi} are considered for the prediction in this paper. 

Some works try to tackle this problem as a spatial-agnostic problem, usually with classic \ac{ml} or \ac{dnn} algorithms. In \cite{Schmid2019a}, the authors train a \ac{lstm} model to predict throughput. To compensate for the lack of position information, low-level mobile network parameters and abstracted location information, e.g., surrounding building size and street type, are used. In \cite{Denizer2024}, an architecture using bi-directional \ac{lstm} is proposed to predict \ac{dl} data-rate for multiple prediction horizons. In \cite{Palaios2023}, the authors investigate the performance as well as the interpretability and explainability of \ac{ml} models systematically and conclude that they can capture the underlying principles of the \ac{qos} without being explicitly programmed, which defends the usage of \ac{ml} and \ac{dnn} models for \ac{pqos}. Furthermore, they show that gradient-boosted decision tree models can even outperform \acp{dnn}. More importantly, the authors discuss the problem of data splitting and data distribution in depth and reveal the fundamental problem of \ac{ml} usage in \ac{pqos}: the models only try to generalize the data distribution based on the training data. This implies that the prediction performance of such trained models can degrade when deployed without explicit retraining in another area due to the difference in data distribution. The phenomenon is often referred to as \textit{concept drift}. To deal with it, the \ac{ml} and \ac{dnn} models need to be retrained with new data, which can be time and resource-consuming. The results in \cite{Schippers2023} confirm the same problem, but disagree upon the generalization capability of the \ac{ml} methods.

The \ac{ml} and \ac{dnn} models excel in capturing the temporal and inter-feature dependency of the data, while downplaying the static spatial features. The spatial dependency is often depicted as a \ac{rem} \cite{Schippers2022, Schippers2023}. Other works, such as \cite{Schmid2018,Schmid2019,Jomrich2018} also use the name \textit{connectivity map}. The \ac{rem} stores the historic data about the mobile network, including \ac{phy} \acp{kpi}, such as \ac{rsrp}, \ac{rsrq}, \ac{sinr}, as well as application layer \acp{kpi}, such as throughput, data-rate, round-trip latency, etc. In \cite{Ain2024}, the authors use the historic data recorded by another vehicle, which was previously driven by the adjacent location, to predict the \ac{qos} for the current vehicle, proving the feasibility of \ac{pqos} using historic data. 

In this paper, we intend to incorporate the two different methods. To achieve that, we first launch a data measurement campaign to generate a dataset. Then, we propose a framework that predicts two important \acp{kpi} of the mobile network for teleoperation: the \ac{ul} data-rate and the round-trip latency. We introduce a method that incorporates the historic data in the form of a \ac{rem} into the prediction framework by training the \ac{ml} model to learn the difference between the target value and the reference values from the historic data, which alleviates the prediction performance degradation due to concept drift without explicit retraining.

This paper is structured as follows: Chapter~\ref{sec:data_measurement} describes the data measurement campaign and the static analysis of the data. Chapter~\ref{sec:qos_prediction} introduces the prediction framework. Chapter~\ref{sec:evaluation} shows the performance of the proposed framework and method based on our dataset. Chapter~\ref{sec:conclusion} concludes the paper and discusses future research directions. 

\section{Data Measurement}
\label{sec:data_measurement}

\subsection{Measurement Setup}
\label{sec:measurement_setup}

To investigate the \ac{qos} of mobile network for teleoperation, we first launch a data measurement campaign to collect data for network condition of teleoperation in real-world, which is made publicly available online \cite{Su2025, Su2025-2}. We use the research vehicle EDGAR for the data measurement campaign \cite{Karle2023}.

\begin{figure}[htbp]
    \centering
    \includegraphics[width=\linewidth]{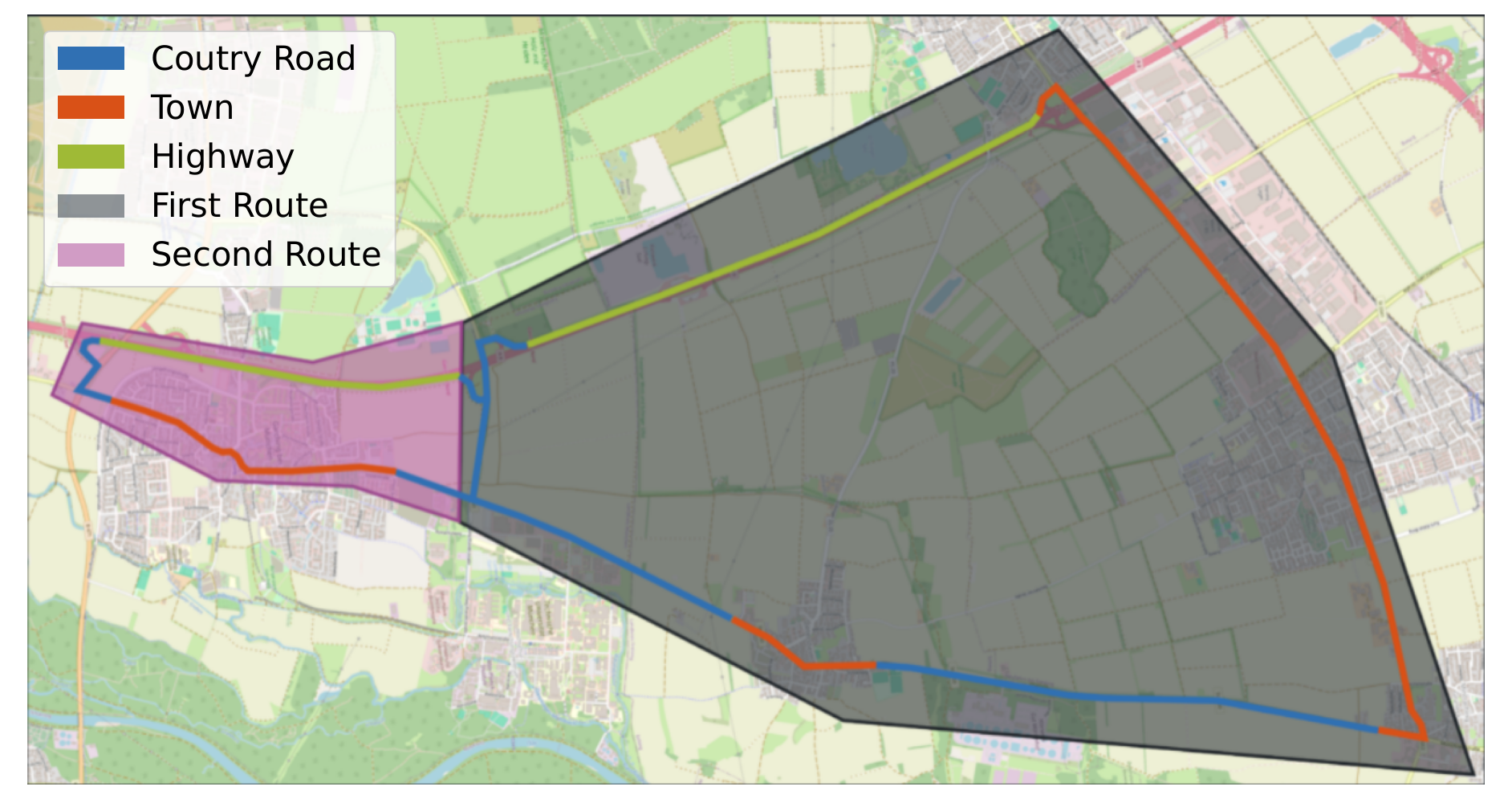}
    \caption{Data measurement routes with different \acp{odd}}
    \label{fig:route_odd}
\end{figure}

For data measurement, we designed two routes that consist of different \acp{odd} for teleoperation, as shown in Fig~\ref{fig:route_odd}. For the training data, we have also taken the temporal factor into consideration. The data are measured at different times on a day and on different days in a week, with a focus on Thursday morning and Sunday afternoon, representing weekdays and weekends. The \ac{ml} model is only trained with data from the first route. The second route represents data outside the area where the training data are collected. The \ac{ml} model is not trained with any data from the second route at all.

The data measurement includes two types of \acp{kpi} of the mobile network of teleoperation, i.e., the \ac{phy} \acp{kpi} and the application layer \acp{kpi}. The \ac{phy} \acp{kpi} are recorded using the Milesight UR75 Industrial 5G Router on board. The application layer \acp{kpi} are recorded between the teleoperated vehicle and the remote operator PC located on campus. The round-trip latency is measured using \textit{ping} by sending an \texttt{icmp} packet. During the latency measurement, a set of constant video streams with 4500 Kbps is sent from the teleoperated vehicle to the remote operator PC to simulate realistic teleoperation. The \ac{ul} data-rate is measured using \textit{iperf3} by sending UDP packets from the teleoperated vehicle to the remote operator PC \cite{iperf3}. The latency and \ac{ul} data-rate are measured in different runs, since the \ac{ul} data-rate measurement will use practically all \ac{ul} bandwidth and therefore render the latency unrealistically high. 

The measured data are synchronized into data frames between different data sources using the timestamps. The data recorded on the first route are also split by runs for the model performance evaluation. The data recorded on the second route are split as historic data to generate a \ac{rem}.

\begin{figure*}[ht]
    \centering
    \includegraphics[width=\textwidth]{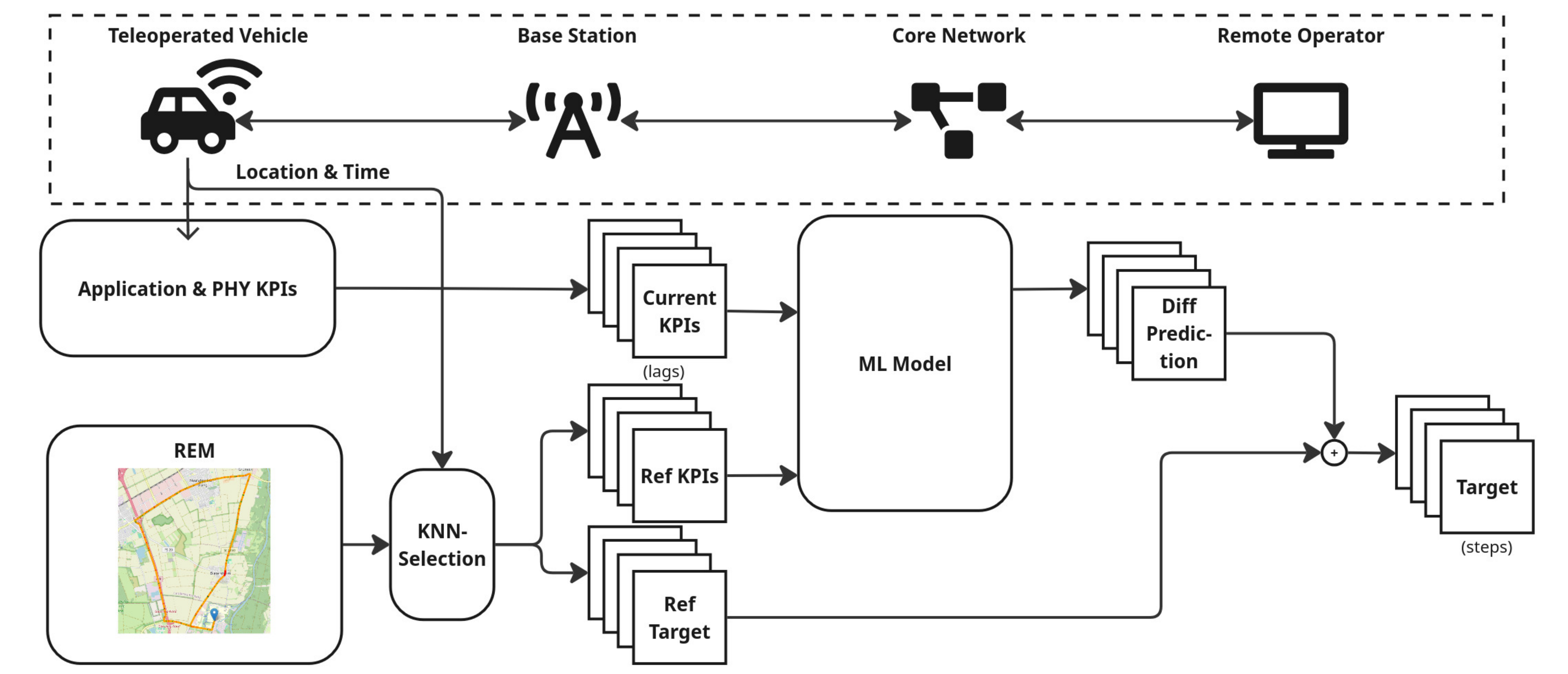}
    \caption{\ac{pqos} framework for teleoperation that uses historic data from \ac{rem} and real-time measurement from teleoperated vehicles}
    \label{fig:prediction_pipeline}
\end{figure*}

\begin{table}[ht]
\centering
\caption{Correlation between Application Layer and \ac{phy} \acp{kpi}}
\begin{tabularx}{\linewidth}{l*{6}{>{\centering\arraybackslash}X}}
\toprule
 & RSRQ & RSRP & SINR & Throughput & Lat. & Long. \\
\midrule
UL & 0.32 & 0.45 & 0.34 & / & 0.23 & 0.15 \\
Latency     & -0.06 & -0.04 & -0.04 & 0.02 & -0.03 & 0.02 \\
\bottomrule
\end{tabularx}
\label{tab:kpi_correlation}
\end{table}

\subsection{Static Data Analysis}
\label{sec:static_data_analysis}


Based on the measured data, we first analyze the correlation between the \ac{phy} \acp{kpi} and the application layer \acp{kpi}. The correlation shown in Tab.~\ref{tab:kpi_correlation} shows that \ac{rsrp}, \ac{rsrq}, and \ac{sinr} show higher correlation to \ac{ul} data-rate. However, our results do not show a monotonically increasing relationship between the three \ac{phy} \acp{kpi} and the \ac{ul} data-rate. It is shown it is unlikely to have low \ac{ul} data-rate with good \ac{rsrp}, \ac{rsrq}, or \ac{sinr}. However, it is not a necessary condition for a high \ac{ul} data-rate. Similar correlation is not observed for latency.

\begin{figure}[htbp]
    \centering
    \includegraphics[width=\linewidth]{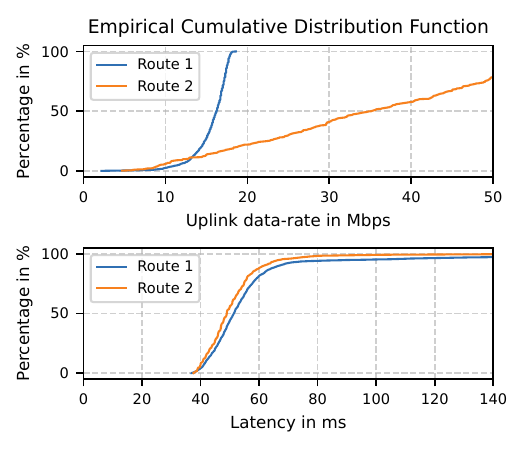}
    \caption{Empirical Cumulative Distribution Function (ECDF) for the UL data-rate and round-trip latency of the two routes}
    \label{fig:ecdf}
\end{figure}

We also compare the \ac{ecdf} of the \ac{ul} data-rate and the latency between the two routes in Fig.~\ref{fig:ecdf}. The results show that the \ac{ul} is more susceptible to changes. This suggests that the \ac{ul} data-rate is more likely to undergo a concept drift. This indicates that the prediction performance of the \ac{ml} model can degrade when deployed on unseen data without explicit retraining.

\section{Quality of Service Prediction}
\label{sec:qos_prediction}

\subsection{Prediction Framework}
\label{sec:prediction_framework}

As an \ac{ott} solution for \ac{pqos}, the prediction framework is deployed on the teleoperated vehicle, as shown in Fig.~\ref{fig:prediction_pipeline}. The current measurements for \ac{phy} \acp{kpi} and application layer \acp{kpi} are taken online on the vehicle and used for prediction. Unlike many recent works, we focus on predicting the \ac{ul} data-rate and the round-trip latency for teleoperation, as our experience suggests that the \ac{ul} is more critical than the \ac{dl} for teleoperation. However, we cannot actually measure the \ac{ul} data-rate during run-time, as the measurement will spam the \ac{ul} and therefore render it unusable for teleoperation. Thus, it is not included as an input feature for prediction.

The main difference of the proposed framework is that the \ac{ml} model is trained to predict the difference between the target values and the reference values from the historic data. The historic data stored in the \ac{rem} is selected by a \ac{knn} algorithm and the reference values are calculated based on the selected historic data. The historic data also contains reference values for the input feature \acp{kpi}, as shown in Tab.~\ref{tab:input_features}. After inference, the predicted difference and the reference values add up to the actual target values. 

\begin{table}[htbp]
\centering
\caption{Input Feature \acp{kpi} for Online Inference}
\begin{tabularx}{\linewidth}{@{}lX@{}}
\toprule
\multicolumn{2}{@{}l}{\textbf{PHY}} \\
\midrule
RSRP  & Reference Signal Received Power - indicates the strength of the received signal from the base station. \\
RSRQ  & Reference Signal Received Quality - reflects the quality of the received reference signal and accounts for interference. \\
SINR  & Signal to Interference plus Noise Ratio - measures the signal quality considering both interference and noise. \\
CQI   & Channel Quality Indicator - measures the quality factors of the radio signal and the radio channel between the end user equipment and the base station \\

\addlinespace[0.5em]
\midrule
\multicolumn{2}{@{}l}{\textbf{Application Layer}} \\
\midrule
Throughput & Actual \ac{ul} throughput that are used by the software on vehicle \\
Latency & Round-trip latency between the teleoperated vehicle and the remote operator measured by \texttt{ping} \\

\bottomrule
\end{tabularx}
\label{tab:input_features}
\end{table}

To capture the temporal relation between different timestamps, we not only use the data measurement for the current timestamp but also the data measurement from the last couple of timestamps. The number of past timestamps are denoted as \texttt{lag}. Considering different use cases, such as video stream bit-rate adaptation due to data-rate and velocity restriction due to high latency, we also predict further into the future. The number of frames that predict into the future is denoted as \texttt{step}. Both \texttt{lag} and \texttt{step} are hyper-parameters of the \ac{ml} model used in the framework and can be fine-tuned according to the actual use case of the prediction.


In our framework, we use a combination of \texttt{XGBRegressor} \cite{Chen2016} and \texttt{MultiOutputRegressor} \cite{Pedregosa2011} as the \ac{ml} model to enable the multi-step and multi-head prediction. Before sending the current measurements to the \ac{ml} model, they are first standardized using mean and standard deviation. In addition, the outliers in the data measurements are eliminated. 

\subsection{Usage of Historic Data}

The historic data are selected based on their relevance to the current prediction location and time, which is represented as a distance function in the \ac{knn}. The distance function is modified to consider both spatial and temporal "distance", as shown in Eq.~\eqref{eq:distance}. The first term considers only the Euclidean distance between the two data measurements using Haversine formula. The second term and the third term both consider the temporal distance between the measurements with a different focus. In \cite{Akselrod2017}, the data measurements suggest a regular pattern of mobile network condition fluctuation on a weekly basis. Therefore, we assume the data measurements follow the same weekly pattern. The second term calculates the time difference between two data measurements within a week, while the third term calculates the week difference. Each distance is normalized and weighted. Based on the distance, the nearest neighbors are then selected based on hyper-parameter \texttt{k} and averaged as the reference. 

\begin{align}
\label{eq:distance}
D(x, y) &= \alpha \cdot \mathcal{N}\big(d_{\text{euclid}}(x, y)\big) + \beta \cdot \mathcal{N}\big(d_{\text{week}}(t_x, t_y)\big) \notag \\
       &\quad + \gamma \cdot \mathcal{N}\big(d_{\text{inweek}}(t_x, t_y)\big) \\
\label{eq:euclid}
d_{\text{euclid}}(x, y) &= \left\| \mathbf{x} - \mathbf{y} \right\|_2 \\
\label{eq:inweek}
d_{\text{inweek}}(t_x, t_y) &= \left| (t_x \bmod s_w) - (t_y \bmod s_w) \right| \\
\label{eq:week}
d_{\text{week}}(t_x, t_y) &= \left| \left\lfloor \frac{t_x}{s_w} \right\rfloor - \left\lfloor \frac{t_y}{s_w} \right\rfloor \right|
\end{align}

\noindent
Where:
\begin{itemize}
  \item \( \alpha, \beta, \gamma \) are the weights for each term.
  \item \( \mathcal{N}(\cdot) \) denotes a normalization function applied to each distance component.
  \item \( \mathbf{x}, \mathbf{y} \) are location represented in latitude and longitude.
  \item \( t_x, t_y \) are timestamps (in seconds).
  \item \( s_w = 604800 \) is the number of seconds in a week.
\end{itemize}

The hyper-parameters used for the \ac{ml} model and the \ac{knn} algorithm are shown in Tab.~\ref{tab:hyperparameters}. The hyper-parameters are chosen through parameter searches.

\begin{table}[htbp]
\centering
\caption{Hyper-parameters for Online Inference}
\begin{tabularx}{\linewidth}{@{}l l X@{}}
\toprule
\textbf{ML Model} & \textbf{Value} & \textbf{Description} \\
\midrule
lag                 & 60                & Number of past data frames included in the prediction framework \\
step                & 60                & Number of time horizons predicted into the future \\
max depth           & 4                 & Maximum tree depth for base learners\\
number of estimators    & 100           & Number of gradient boosted trees\\
\midrule
\textbf{REM} & \textbf{Value} & \textbf{Description} \\
\midrule
distance weights    & [0.8, 0.2, 0.6]   & The weight vector for the three distances in \ac{knn} algorithm \\
k                   & 20                & Number of nearest neighbors selected in \ac{knn} algorithm\\
\bottomrule
\end{tabularx}
\label{tab:hyperparameters}
\end{table}

\begin{figure}[h]
    \centering
    \includegraphics[width=\linewidth]{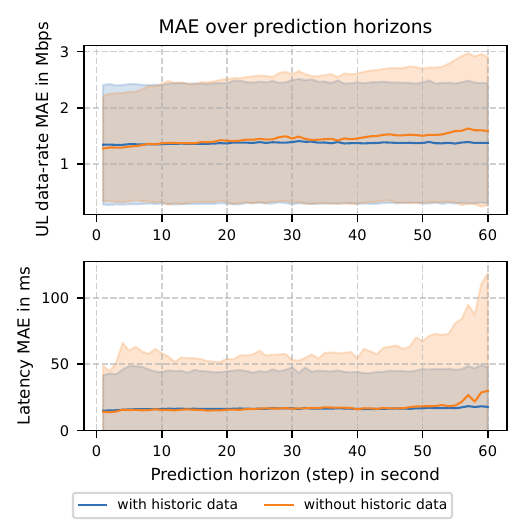}
    \caption{Mean absolute error (MAE) for uplink data-rate and round-trip latency prediction over prediction horizons for route 1}
    \label{fig:mae}
\end{figure}

\begin{figure*}[htbp]
    \centering
    \includegraphics[width=\linewidth]{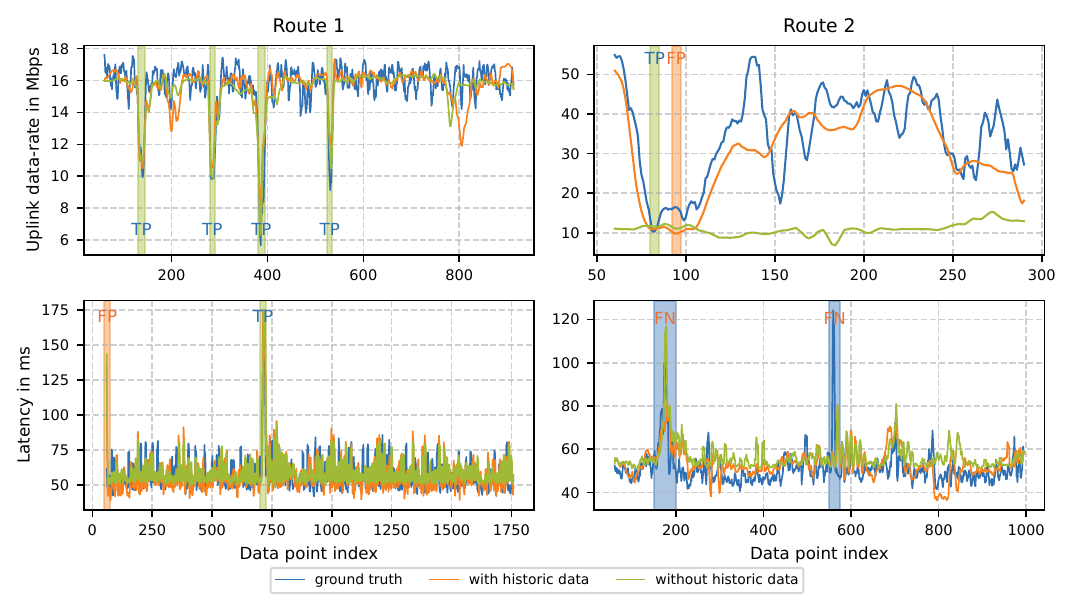}
    \caption{Comparison of \ac{ul} data-rate and round-trip latency prediction between with historic data and without historic data for route 1 (same route as where the training data are recorded) and route 2 (completely unseen data). The prediction horizon is for the next step, i.e., second. The data are smoothed with a window of 5 seconds and the outliers are removed for better visualization. \textit{True positive} (\textbf{TP}), \textit{false positive} (\textbf{FP}), and \textit{false negative} (\textbf{FN}) showcase the performance of critical scenario detection.}
    \label{fig:ul_pred_comparison}
\end{figure*}

\section{RESULTS}
\label{sec:evaluation}


\subsection{General Prediction Performance}
\label{sec:prediction_performance}

To demonstrate the problem with traditional prediction pipeline, we trained the same \ac{ml} model to directly predict \ac{ul} data-rate and round-trip latency. The data recorded on the first route are split into training data and testing data based on measurement runs. The data recorded on the second route are not included in the training data. The results shown in Fig.~\ref{fig:ul_pred_comparison} show the traditional prediction model without using historic data can predict well for the first route, but its performance quickly degrades on the unseen data in the second route, since the distribution of the unseen data varies drastically from its training data. To cope with this, the model usually needs retraining, which is time and resource consuming. 

However, by using historic data as reference points, the prediction of the \ac{ml} model can be greatly improved for unseen data, while maintaining similar performance for the first route. The results indicate that the \ac{ml} model can learn the difference between the target values and the reference values based on the current and reference \acp{kpi}. It is worth noticing that the prediction performance of proposed framework does rely on the accuracy of the historic data, but the model does not necessarily have to undergo retraining as the improvement on the accuracy of the historic data will also in turn improve the performance of the prediction. 

A slight degradation for higher prediction horizons for the first route is observed in Fig~\ref{fig:mae}. By using historic data, the prediction error reduces for higher prediction horizons, as the historic data act as anchor point for the prediction. A similar relationship for the second route is however not observed.

\begin{table}[htbp]
\centering
\caption{F1, precision, and recall for critical scenario detection for teleoperation}
\begin{tabularx}{\linewidth}{@{}l l c c c@{}}
\toprule
\textbf{Route 1} & \textbf{Condition} & \textbf{F1} & \textbf{Precision} & \textbf{Recall} \\
\midrule
UL with historic data            & $<$14 Mbps    & 0.467 & 0.549 & 0.406 \\
UL without          & $<$14 Mbps    & \textbf{0.505} & 0.688 & 0.399 \\
Latency with historic data       & $>$100 ms     & \textbf{0.503} & 0.465 & 0.548 \\
Latency without       & $>$100 ms     & 0.471 & 0.456 & 0.488 \\
\midrule
\textbf{Route 2} & \textbf{Condition} & \textbf{F1} & \textbf{Precision} & \textbf{Recall} \\
\midrule
UL with historic data            & $<$14 Mbps    & \textbf{0.489} & 0.333 & 0.917 \\
UL without            & $<$14 Mbps    & 0.102 & 0.054 & 1.000 \\
Latency with historic data       & $>$100 ms     & \textbf{0.167} & 0.500 & 0.100 \\
Latency without       & $>$100 ms     & 0.000 & 0.000 & 0.000 \\
\bottomrule
\end{tabularx}
\label{tab:f1results}
\end{table}

\subsection{Critical Scenario Detection for Teleoperation}
\label{sec:critical_scenario}

For teleoperation, the general prediction performance does not fully represent its use case for \ac{pqos}, since minor network fluctuation does not affect the teleoperation. It will only become critical, when the data-rate and the latency surpass certain thresholds. Therefore, we introduce the metric for detecting critical scenarios to evaluate the prediction performance specifically for teleoperation, as demonstrated in Fig.~\ref{fig:ul_pred_comparison}. In this paper, we take the emprical values of 14 Mbps for \ac{ul} data-rate and 100 ms for round-trip latency as thresholds, as we notice that the performance of teleoperation tends to become instable under or above such values. Furthermore, we calculate the F1 score to evaluate the performance by taking both \textit{False Negative} and \textit{False Positive} into consideration, as shown in Tab.~\ref{tab:f1results}. For the first route, the F1 scores are slightly higher for \ac{ul} data-rate without the historic data, but actually lower for latency. For the unseen data in the second route, F1 scores are higher for both \ac{ul} data-rate and latency with historic data. The results also show the difficulty in predicting latency for unseen data in general.

\section{CONCLUSION}
\label{sec:conclusion}

In this paper, we first launch a data measurement campaign to record the \ac{qos}-related \acp{kpi} of commercial mobile network with an experimental teleoperated vehicle, including \ac{phy} \acp{kpi} and application layer \acp{kpi}. Based on the data, we conduct a static analysis between the two routes used for evaluation. Then, as an \ac{ott} solution to \ac{pqos}, we propose a framework to predict two crucial \acp{kpi} of teleoperation, i.e., \ac{ul} data-rate and round-trip latency. A method is introduced to incorporated historic data measurements into the online prediction pipeline by training the \ac{ml} model to predict the difference to the reference target values from the historic data. This reduces the prediction error on previously unseen data due to concept drift. At last, we evaluate the prediction performance using mean absolute error. The results show that the method is able to reduce the prediction error compared to the traditional approach. In addition, we evaluate the ability of proposed \ac{pqos} framework to detect critical scenarios in teleoperation. We show that the  framework is able to detect more \ac{ul} data-rate related critical scenarios. 

However, stemming from results of this work, some questions still need to be answered for this topic. First, the proposed approach needs to be evaluated on more routes and devices with a wider range of data distribution, as the historic data or \ac{rem} are likely to be shared between different end devices. Second, the concrete usage of the \ac{qos} prediction for teleoperation needs to be addressed. More specifically, the counter-measurement of a teleoperation software to critical network scenarios need to be defined and investigated.

\addtolength{\textheight}{-12cm}   

\begin{acronym}
    \acro{pqos}[pQoS]{Predictive Quality of Service}
    \acro{qos}[QoS]{Quality of Service}
    \acro{ad}[AD]{Autonomous Driving}
    \acro{mno}[MNO]{Mobile Network Operator}
    \acro{ott}[OTT]{Over-the-Top}
    \acro{ran}[RAN]{Radio Access Network}
    \acro{lte}[LTE]{Long-Term Evolution}
    \acro{dl}[DL]{Downlink}
    \acro{ul}[UL]{Uplink}
    \acro{kpi}[KPI]{Key Perfomrance Indicator}
    \acro{rem}[REM]{Radio Environmental Map}
    \acro{knn}[KNN]{K-Nearst-Neighbor}
    \acro{dnn}[DNN]{Deep Neural Network}
    \acro{lstm}[LSTM]{Long Short-Term Memory}
    \acro{odd}[ODD]{Operational Design Domain}
    \acro{phy}[PHY]{Physical Layer}
    \acro{ml}[ML]{Machine Learning}
    \acro{ecdf}[ECDF]{Empirical Cumulative Distribution Function}
    \acro{rsrp}[RSRP]{Reference Signal Received Power}
    \acro{rsrq}[RSRQ]{Reference Signal Received Quality}
    \acro{sinr}[SINR]{Signal-Interference-to-Noise-Ratio}

    \acro{3gpp}[3GPP]{3rd Generation Partnership Project}
    \acro{5gaa}[5GAA]{5G Automotive Association}

    \acro{mae}[MAE]{Mean Absolute Error}
    \acro{rmse}[RMSE]{Root Mean Square Deviation}
    
\end{acronym}
\section*{ACKNOWLEDGMENT}
\label{sec:acknowledgement}

Xiyan Su is the first author of the paper and main contributor to the presented work. As second authors, Jianning Gao and Mahmoud Ashri contributed to the development and implementation of the \ac{ml} model and historic data approach respectively and revised the paper critically. Frank Diermeyer made contributions to the conception of the research project and revised the paper critically.

\bibliographystyle{IEEEtran}
\bibliography{bib}

\end{document}